\documentclass[journal=jacsat,manuscript=article]{achemso}

\mciteErrorOnUnknownfalse

\usepackage[utf8]{inputenc} 
\usepackage[T1]{fontenc}    
\usepackage{hyperref}       
\usepackage{url}            
\usepackage{booktabs}       
\usepackage{amsfonts}       
\usepackage{nicefrac}       
\usepackage{microtype}      
\usepackage{xcolor}         
\usepackage{graphicx}

\title{Interpretable Deep Learning for Polar Mechanistic Reaction Prediction}

\author{Ryan J. Miller}
\email{rjmille3@uci.edu}
\affiliation[csdep]
{Department of Computer Science,
University of California, Irvine, Irvine, California 92697, United States}

\author{Alexander E. Dashuta}
\affiliation[chemdep]
{Department of Chemistry,
University of California, Irvine, Irvine, California 92697, United States}

\author{Brayden Rudisill}
\affiliation[csdep]
{Department of Computer Science,
University of California, Irvine, Irvine, California 92697, United States}

\author{David Van Vranken}
\email{david.vv@uci.edu}
\affiliation[chemdep]
{Department of Chemistry, University of California, Irvine, Irvine, California 92697, United States}

\author{Pierre Baldi}
\email{pfbaldi@uci.edu}
\affiliation[csdep]
{Department of Computer Science,
University of California, Irvine, Irvine, California 92697, United States}

\begin{document}

\maketitle

\begin{abstract}

Accurately predicting chemical reactions is essential for driving innovation in synthetic chemistry, with broad applications in medicine, manufacturing, and agriculture. At the same time, reaction prediction is a complex problem which can be both time-consuming and resource-intensive for chemists to solve. Deep learning methods offer an appealing solution by enabling high-throughput reaction prediction. However, many existing models are trained on the US Patent Office dataset and treat reactions as overall transformations—mapping reactants directly to products with limited interpretability or mechanistic insight. To address this, we introduce PMechRP (Polar Mechanistic Reaction Predictor), a system that trains machine learning models on the PMechDB dataset, which represents reactions as polar elementary steps that capture electron flow and mechanistic detail. To further expand model coverage and improve generalization, we augment PMechDB with a diverse set of combinatorially generated reactions. We train and compare a range of machine learning models, including transformer-based, graph-based, and two-step siamese architectures. Our best-performing approach was a hybrid model, which combines a 5-ensemble of Chemformer models with a two-step Siamese framework to leverage the accuracy of transformer architectures, while filtering away "alchemical" products using the two-step network predictions. For evaluation, we use a test split of the PMechDB dataset and additionally curate a human benchmark dataset consisting of complete mechanistic pathways extracted from an organic chemistry textbook. Our hybrid model achieves a top-10 accuracy of 94.9\% on the PMechDB test set and a target recovery rate of 84.9\% on the pathway dataset.

\end{abstract}

\section{Introduction}

Three main approaches exist for the prediction of chemical reactions: quantum chemistry based methods \cite{balabin2009neural, pinheiro2020machine, curtiss2007gaussian, kadish2021methyl}, rule based methods \cite{chen2009no}, and machine learning (ML) based methods \cite{schwaller2021prediction, probst2022reaction, coley2019graph, coley2017convolutional, fooshee2018deep, zheng2019predicting, coley2017prediction, kayala2012reactionpredictor, bradshaw2018generative, segler2017neural, joung2024reproducing}. Quantum chemistry methods offer highly accurate predictions of chemical properties, but their significant computational cost renders them slow and limits their use for broad, high-throughput reaction prediction. On the other end of the spectrum, rule-based models offer rapid predictions, but suffer from inflexibility. Because chemical reactions span an infinite and extremely complex space, encoding them into a fixed set of rules is inherently limiting. Such systems often fail when they encounter reactions outside their predefined scope. For a balance between precision and speed, ML models offer both flexibility and scalability, making them well-suited for application across larger chemical systems and datasets. Countless ML models have been devised for tasks such as reaction yield prediction \cite{schwaller2021prediction}, reaction classification \cite{probst2022reaction}, chemical property prediction \cite{coley2019graph, coley2017convolutional}, and both forward and reverse reaction prediction \cite{fooshee2018deep, zheng2019predicting, coley2017prediction, kayala2012reactionpredictor, bradshaw2018generative, segler2017neural, joung2024reproducing}.

\begin{figure}
    \centering
\includegraphics[width=1.0\linewidth]{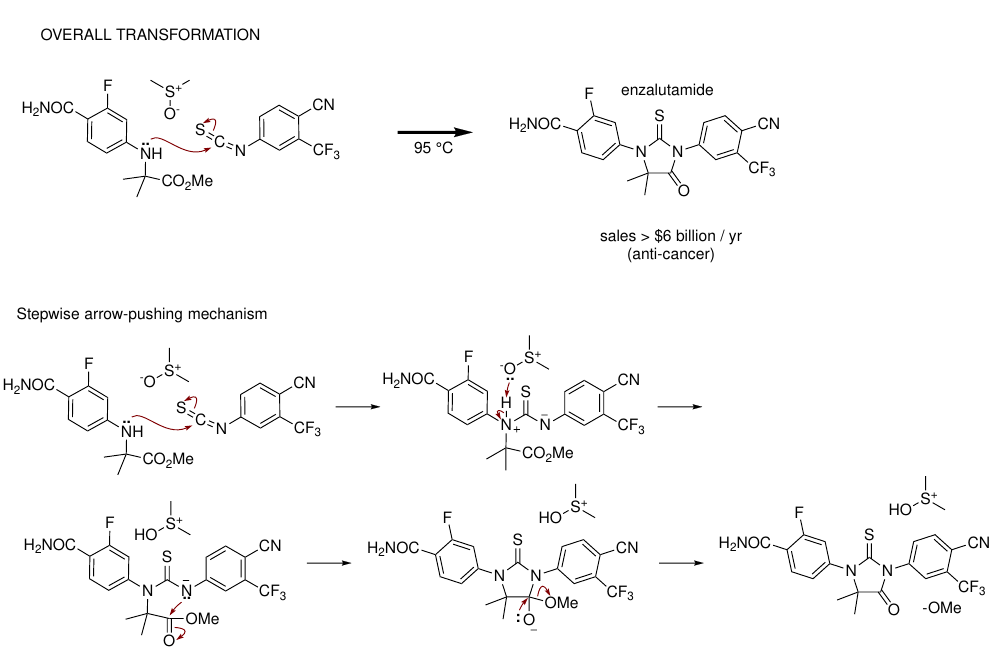}
    \caption{Example of an overall transformation vs an elementary step approach. This is a the final reaction step in the synthesis of enzalutamide, a drug used to treat prostate cancer that generates over \$6 billion a year in revenue \cite{zhou2017improved}.}
    \label{fig:overall}
\end{figure}

Although ML models offer high-throughput and highly adaptable chemical prediction, a significant drawback lies in their lack of interpretability in comparison to quantum chemistry based methods. The predominant approach of training models on the USPTO (US Patent Office) dataset \cite{lowe2012extraction}, means many ML models predict reactions as overall transformations. This results in a black-box scenario, where predicted products emerge directly from reactants without insight into intermediate transition states. Although these models may achieve high accuracy on the USPTO dataset, their outputs pose challenges for organic chemists, who typically reason through chemical synthesis via arrow-pushing mechanisms rather than overall transformations. An example of the overall transformation versus a mechanistic elementary step approach can be seen in Figure \ref{fig:overall}. The elementary step approach breaks the overall transformation down into a sequence of arrow-pushing steps, which illustrate the flow of electrons and the shifting of atoms.

\begin{figure}
    \centering
    \includegraphics[width=1.0\linewidth]{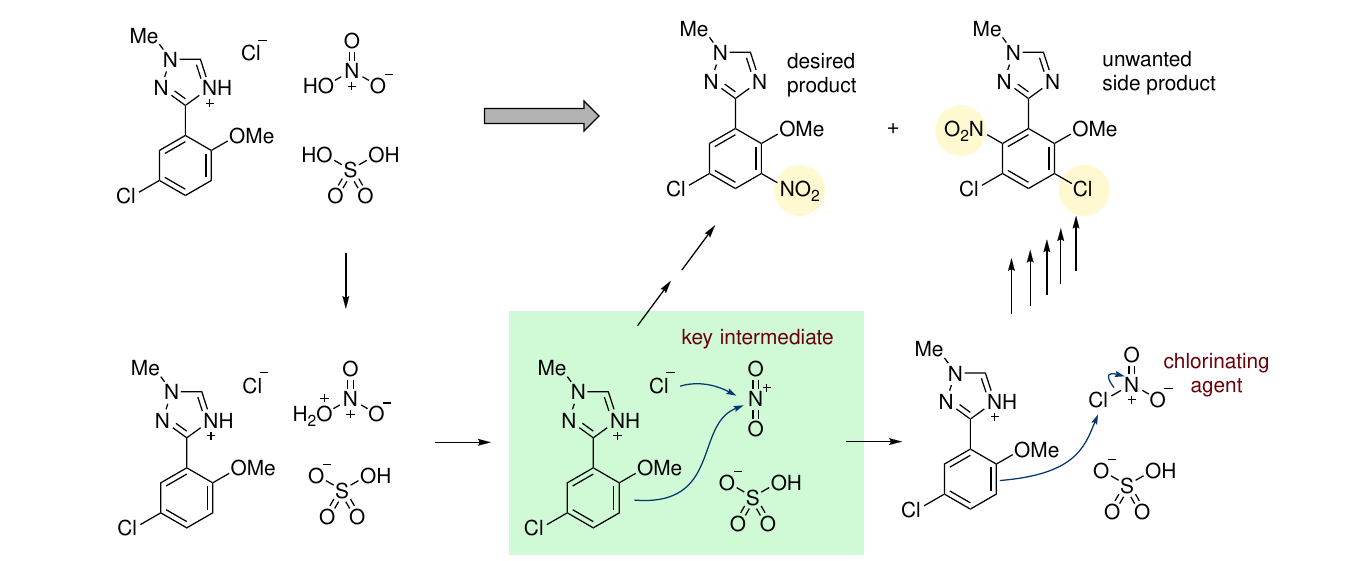}
    \caption{A side reaction occurring at an intermediate step in the synthesis of the autoimmune drug Deucravacitinib, generated unwanted side products due to competing addition of chloride anion to the key NO$_{2}^+$ intermediate. This led to a decrease in overall purity of the products \cite{treitler2022development}.}
    \label{fig:importance_of_elementary}
\end{figure}

By thinking of reactions as occurring through elementary steps, organic chemists can reason about the underlying driving forces of a reaction. These mechanistic insights help explain phenomena such as unexpected side products or variations in product yield. Figure \ref{fig:importance_of_elementary} illustrates the importance of understanding these intermediate steps in a mechanistic pathway where the purity of the final products was affected by a side reaction. When training ML models to forecast elementary step reactions, we effectively guide them to emulate organic chemists' thought processes, thereby generating predictions that are more easily interpretable and serve as practical aids for organic synthesis design.

A further limitation of the popular USPTO dataset is the presence of a substantial number of unbalanced reactions. Training on such reactions can lead to models that produce unbalanced predictions, which poses particular problems for pathway prediction. When expanding the tree of plausible reactions during a pathway search, it is critical that all atoms are accounted for in each step—otherwise, the predicted pathways may "lose" atoms, creating branches which do not have access to all available reactive atoms. In contrast, datasets like PMechDB, which are both balanced and mechanistically annotated, provide a more chemically rigorous foundation for model training.

\section{Data}

\subsection{Manually Curated}

To develop predictive models for polar reaction mechanisms, we trained on the recently introduced PMechDB dataset. This dataset consists of approximately 13,000 polar elementary steps, each balanced, partially atom mapped, and manually verified by a team of organic chemists. Each reaction represents a single elementary step polar reaction. These entries have been collected through manual curation from a diverse array of chemistry literature and textbooks \cite{tavakoli2024pmechdb}.

\subsection{Combinatorial Reactions}

In addition to the manually curated data, we also utilized a novel dataset consisting of approximately 48 million combinatorially generated proton transfer reactions \cite{Dashuta2025ptdataset}. These reactions are generated by pairing acids and bases together and calculating rate constants to filter away implausible reactions. More details on the combinatorial reaction generation can be found in the Appendix.

\subsection{Human Benchmark Pathway Dataset}

Lastly, we curated a dataset of 350 mechanistic pathways from an intermediate-level organic chemistry textbook. Each pathway consists of a set of reactants, a target product, and a sequence of 1 to 7 plausible mechanistic steps. To establish a human benchmark, these pathways were assigned to upper-division chemistry students, who were asked to predict the final product based on the provided reactants. Submissions that were left blank or showed no clear mechanistic reasoning were excluded from evaluation. This filtering resulted in 289 problems, of which 180 were answered correctly—yielding an undergraduate (UG) benchmark accuracy of 62.3\%. Since omitted submissions may reflect an inability to predict the mechanism, this figure should be viewed as a generous estimate of UG performance. Additional details on the curation and evaluation process are provided in the Appendix.

\subsection{Training and Testing Splits}

For the PMechDB data, we perform an 80/10/10 train/val/test split via random sampling. We refer to this as a manually curated data split. We then augment this data split by adding 10k combinatorially generated reactions into the training set. We refer to this augmented data split as the mixed data split. We train and assess all models on both the manually curated and mixed data splits. Lastly, we assess the performance of the best performing model on the human benchmark pathway dataset.

\section{Methods}

Here we describe several different machine learning approaches for predicting polar elementary step mechanisms. These methods fall into two distinct categories: the reactive atom two-step approach, and the single-step seq-to-seq or graph-to-seq prediction methods. 

\subsection{Single-Step Prediction}
We evaluated several transformer-based and graph-based models that treat reaction prediction as either a single-step sequence-to-sequence or graph-to-sequence translation problem, mapping reactant SMILES strings to product SMILES. These include Molecular Transformer \cite{schwaller2019molecular}, Chemformer \cite{irwin2022chemformer}, T5Chem \cite{lu2022unified}, and Graph2SMILES \cite{tu2022permutation}. While these models have demonstrated strong performance on benchmark datasets like USPTO, they do not provide arrow pushing information, fail to enforce chemical validity, and in the case of sequence-to-sequence models, lack permutation invariance. The Appendix contains additional details regarding the training of each model.

\subsection{Two-Step Prediction}
In contrast to black-box single-step models, we implemented a two-step architecture \cite{fooshee2018deep} that explicitly models electron flow via reactive atom identification and arrow-pushing mechanism enumeration. The model first predicts source (electron-donating) and sink (electron-accepting) atoms using dedicated classifiers trained on atom-level features. These predicted sites are then used to generate possible mechanisms via OrbChain \cite{fooshee2018deep, kayala2011learning, tavakoli2024ai}, which are ranked using a Siamese network plausibility model. This approach provides easily interpretable predictions with mechanistic rationale for each step. Additional details describing this methodology are provided in the Appendix.

\subsection{Hybrid Approach}

Drawing from the strengths of both single-step and two-step prediction methods, we propose a hybrid approach that integrates the predictive strength of a 5-ensemble of Chemformer models with the mechanistic validity of the two-step model. While the Chemformer ensemble yields strong predictive performance, it and other transformer-based models are prone to generating "alchemical" products—those with unbalanced charges or atom counts compared to the reactants. To address this, we apply a post-processing filter that identifies and discards chemically invalid predictions. For each reaction, if any ensemble-generated product violates charge or atom conservation, it is replaced by the top-ranked prediction from the two-step model. Because the two-step architecture is grounded in explicit arrow-pushing mechanisms, it ensures mechanistic plausibility. As a result, the final hybrid predictions are now sanity checked for "alchemical" products.

\section{Results and Discussion}

\subsection{Performance on Manually Curated Dataset}

We train all models on the manually curated dataset. The results comparing the performance of the trained models on the test split can be seen in Table \ref{tab:acc_comparison}.

\begin{table}[htbp]
    \centering
    \caption{Top-N Accuracy of Trained Models}
    \label{tab:acc_comparison}
    \begin{tabular}{lcccc} 
        \toprule 
        \textbf{Model Type} & \textbf{Top-1} & \textbf{Top-3} & \textbf{Top-5} & \textbf{Top-10}\\ 
        \midrule 
        Best Two-Step Siamese & 39.5 & 59.6 & 68.2 & 76.8 \\ 
        MolTransformer & 50.2 & 61.3 & 64.5 & 64.5 \\
        T5Chem & 64.1 & 75.9 & 78.4 & 80.3 \\
        Graph2Smiles & 68.3 & 78.9 & 80.8 & 82.8 \\
        Chemformer  & 79.4 & 87.3 & 87.5 & 87.6 \\
        5-Ensemble Chemformer & 81.8 & 90.5 & 91.4 & 91.5 \\
        Hybrid & 81.8 & 91.8 & 93.1 & 94.5 \\
        \bottomrule 
    \end{tabular}
\end{table}

Although the Siamese two-step model allows for improved interpretability due to its direct prediction of arrows, Chemformer yielded the most accurate predictions among all non-ensemble models. Performance of the Chemformer model is significantly improved through ensembling, with further gains achieved by integrating it with the two-step model in a hybrid approach. The hybrid model demonstrates superior performance, achieving a top-10 accuracy of 94.5\%.

\subsection{Performance on Mixed Dataset}

Lastly, we assess the performance benefits of the combinatorial reactions by training several models on the mixed dataset. The accuracy results can be seen in Table \ref{tab:acc_comparison_mixed}.

\begin{table}[htbp]
    \centering
    \caption{Top-N Accuracy of Trained Models on Mixed Dataset}
    \label{tab:acc_comparison_mixed}
    \begin{tabular}{lcccc} 
        \toprule 
        \textbf{Model Type} & \textbf{Top-1} & \textbf{Top-3} & \textbf{Top-5} & \textbf{Top-10}\\ 
        \midrule 
        Best Two-Step Siamese & 36.1 & 55.1 & 63.2 & 72.1 \\ 
        MolTransformer & 53.1 & 65.5 & 67.6 & 67.9 \\ 
        T5Chem & 64.9 & 75.2 & 78.7 & 81.2 \\
        Graph2Smiles & 65.8 & 78.1 & 80.3 & 83.0 \\
        Chemformer  & 79.4 & 87.8 & 88.0 & 88.2 \\
        5-Ensemble Chemformer & 82.1 & 91.1 & 91.8 & 91.9 \\
        Hybrid & 82.0 & 91.9 & 93.6 & 94.9 \\
        \bottomrule 
    \end{tabular}
\end{table}

The addition of combinatorial reactions led to a modest increase in top-5 and top-10 prediction accuracy across most models. Notably, the top-performing Hybrid model saw a 0.5\% improvement in top-5 accuracy, while largest improvement was the MolTransformer model which had its top-5 accuracy improve by over 3\%. However, this trend was not universal: the two-step model experienced a decrease in performance for both top-5 and top-10 accuracy, and the T5Chem model showed a slight decrease in top-5 accuracy, though its top-10 accuracy improved. A possible explanation for the overall performance gains is that the added combinatorial reactions expand the diversity of possible reactants and products, helping models generalize better and reducing overfitting to the relatively small training set of approximately 10,000 manually curated reactions.

\subsection{Pathway Search}
 
We took the human benchmark pathway dataset of 350 mechanistic pathways (containing  reatants, targets, and intermediate steps) with sizes between 1-7 steps and evaluated the performance of the best-performing hybrid model. To predict pathways, we chained the predicted elementary steps together starting from the reactants. We perform a breadth-first search by taking the top 10 predictions from the model. We stop when the time limit exceeds 2 hours or if the target structure is found. In order to speed up the search process, if the model predicts one of the pathway intermediates, including an alternative resonance structure, we immediately branch on this step. We present the results from the model trained on both the manually curated and the mixed datasets in Table \ref{tab:targets_recovered}.

\begin{table}[h!]
    \centering
    \hspace*{-\leftmargin}\begin{tabular}{|c|c|c|c|}
        \hline
        Depth & Total Pathways & Targets Recovered w/o 
combinatorial& Targets Recovered w/ combinatorial\\ \hline
        1 & 37  & 32 & 33 \\ \hline
        2 & 113 & 98 & 99 \\ \hline
        3 & 108 & 98 & 97 \\ \hline
        4 & 35  & 24 & 23 \\ \hline
        5 & 38  & 30 & 31 \\ \hline
        6 & 16  & 9 & 11 \\ \hline
        7 &  3 & 3 & 3 \\ \hline
        all &  350 & 294 & 297 \\ \hline
    \end{tabular}
    \caption{Targets Recovered at Different Depths.}
    \label{tab:targets_recovered}
\end{table} 

The hybrid model without combinatorial reactions recovered the target 84.0\% of the time, while the hybrid model with added combinatorial reactions was able to recover the target 84.9\% of the time. This is a clear improvement over the 62.3\% accuracy obtained by students in the UG benchmark.

The system was effective at recovering target molecules based on molecular formulae, which are often obtainable from high-resolution mass spectrometry. To conserve computational resources, we halted pathway searches after identifying the first route that matched the target. To further assess model performance, our team of trained organic chemists manually reviewed each predicted pathway and evaluated its chemical plausibility. The results can be seen in table \ref{tab:plausibility_targets_recovered}.

\begin{table}[h!]
    \centering
    \begin{tabular}{|c|c|c|c|}
        \hline
        Depth & \# 1st Path to Target & Plausible Pathways & \% Plausible \\ \hline
        1 & 33& 27& 82\\ \hline
        2 & 99& 69& 70\\ \hline
        3 & 97& 45& 46\\ \hline
        4 & 23& 9& 39\\ \hline
        5 & 31& 7& 23\\ \hline
        6 & 11& 2& 18\\ \hline
        7 & 3& 1& 33\\ \hline
    \end{tabular}
    \caption{Plausibility of First Pathways to Target Found.}
    \label{tab:plausibility_targets_recovered}
\end{table}

Focusing only on the first recovered pathway, we observed that overall plausibility decreased with increasing pathway length—from 82\% for 1-step pathways to just 18\% for the 6-step pathways. Some common implausible processes in these pathways involved two-step displacements depicted as one-step displacements, two-step $S_N1$ processes, and two-step addition-elimination at acyl groups, silicon, tin, sulfonyl groups, etc. In many cases, the longer two-step pathways involved high-scoring steps. For example, a two-step nucleophilic aromatic substitution of 1-chloro-4-nitrobenzene by hydroxide anion was incorrectly predicted as a one-step pathway with a step score of 0.08. The true pathway involves two steps: addition (score 1.198) and elimination (score 3.661). However, because the one-step pathway found the target sooner, the pathway search was terminated. One possible improvement would be to run the pathway search for a large amount of time to find as many pathways as possible, and then only show the pathway which contains the maximum low-scoring step, akin to the principle that reactions proceed through the pathway with the fastest rate determining step.

\begin{figure}
    \centering
    \includegraphics[scale = 0.8]{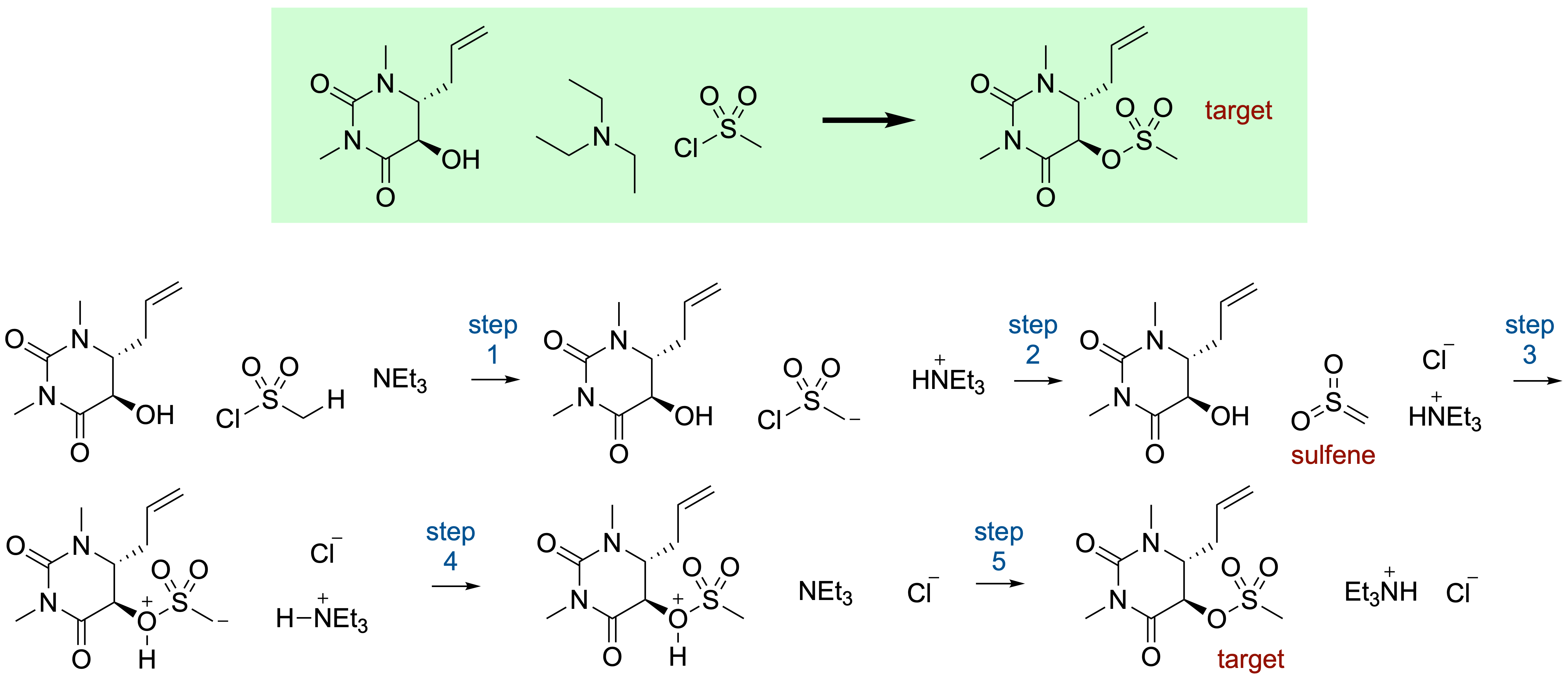}
    \caption{The first pathway generated for mesylation of an alcohol with methanesulfonyl chloride.}
    \label{fig:mesylation}
\end{figure}

Encouragingly, the system trained with additional proton transfers was able to recover more nuanced pathways that included uncommon intermediates. For example, in the mesylation of a complex alcohol to produce $C_{10}H_{16}N_{2}O_{5}S$, the model correctly predicted both the target structure and a chemically plausible pathway that proceeds through a rare sulfene intermediate (Figure \ref{fig:mesylation}).

\begin{figure}
    \centering
    \includegraphics[scale = 0.1]{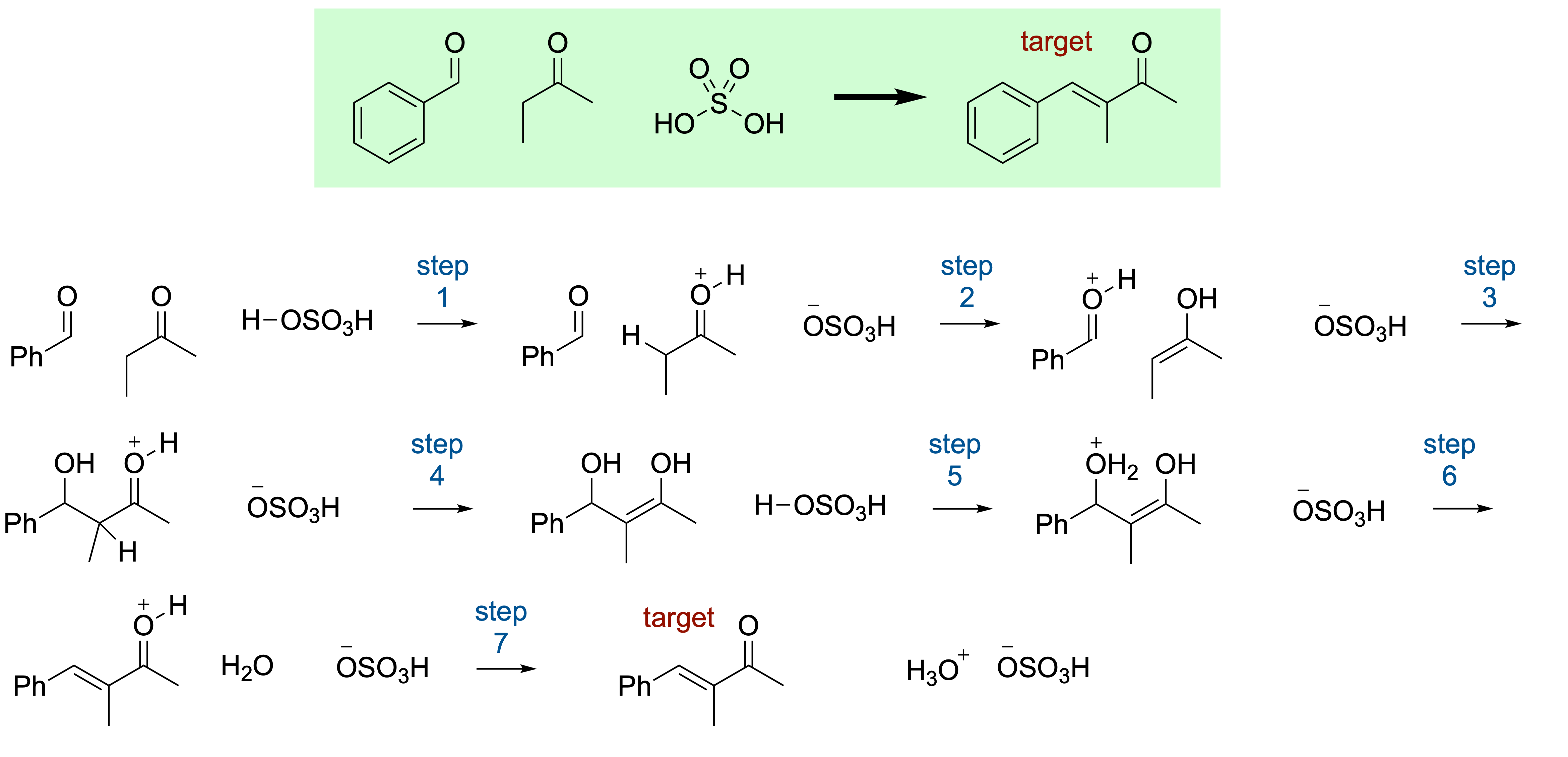}
\caption{The first pathway generated for an aldol condensation under acidic conditions. As with most reactions involving proton transfers, other mechanistic variations are plausible. For example, the formation of the oxonium and enol in step 2 could have been depicted as a two-step process (e.g., step 2a and 2b, not shown) using the bisulfate anion to affect the proton transfer.}
    \label{fig:cropacidaldol}
\end{figure}

While aldol reactions are most commonly performed under basic conditions, our model accurately predicted the acid‑catalyzed aldol condensation in Figure \ref{fig:cropacidaldol}, correctly recovering the expected seven‑step mechanism\cite{Too2011acidaldol}. Although Jung et al.’s template‑based dataset includes base‑catalyzed aldol steps, it excludes acid‑catalyzed reaction templates \cite{Jung2011templates}, so the reaction in Figure \ref{fig:cropacidaldol} would fall outside of its coverage. Template‑based methods offer significant scalability, but their dependence on predefined patterns not only constrains the mechanistic diversity they can capture, but also biases models toward the most common templates—potentially overlooking less frequent yet chemically valid mechanistic pathways.

\subsection{PMechRP Web Interface}

We make the Hybrid, Ensemble Chemformer, and Two-Step models publicly available through an interactive online interface at https://deeprxn.ics.uci.edu/pmechrp. The software offers two options: single-step prediction and pathway prediction. For single-step prediction, the user inputs a set of reactants and model parameters, and the website will offer top-N predictions of the elementary-step mechanism. For pathway prediction, the user inputs a set of reactants, reaction conditions, and a target. The website will then perform a pathway search to find a multistep mechanism which leads from the reactants to the desired product. Parameters such as the branching factor and search depth can be tuned from this interface. Lastly, all training datasets are available under the name "PMechRP" at https://deeprxn.ics.uci.edu/pmechdb/download.

\section{Limitations}

We note there are several limitations with the current state of the PMechRP polar reaction system. First, the PMechDB dataset only includes around 13,000 steps. This means the dataset is relatively small for training large architectures, and it may be difficult for these models to generalize well to all forms of experimental chemistry. To improve coverage, we augment the dataset with combinatorially generated reactions. However, these additional reactions are constructed from a limited set of acids and bases, and while helpful, they do not capture the diversity of chemical space. As such, the overall dataset remains limited in scope compared to the complexity of real-world chemistry. Secondly, the transformer-based models directly translate from the reactants to products, without generating the arrow-pushing mechanisms. Although the elementary step predictions still offer significant interpretability, using methods such as the two-step Siamese method would offer greater insight into the causality of a reaction by directly showing the flow of electrons. Additional methods could be developed to predict arrow codes or reactive orbitals using a transformer architecture in order to offer predictions with arrow-pushing mechanisms. Lastly, by performing hybrid or ensembling methods, the best-performing models have increased computational overhead, and the inference time is comparatively slow.

\section{Conclusion}

We developed and compared several reaction prediction systems for polar reaction mechanisms, demonstrated performance benefits when using combinatorially generated reactions to augment the training set, and introduced a novel dataset of mechanistic pathways for benchmarking elementary step prediction. Through our analysis, we created PMechRP—a reaction prediction system that specifically targets polar reactions at the mechanistic level. Our hybrid method, which combines the Chemformer and two-step architectures, achieves a 94.9 \% top-10 accuracy on the PMechDB test set, and an 84.9\% target recovery rate on the pathway dataset. Together, these contributions represent a step toward more interpretable and mechanism-aware reaction prediction systems.

\bibliography{citations}

\newpage

\appendix
\section{Appendix}

In this appendix, we provide additional details about the data sets, experiments, and models trained.

\subsection{Combinatorial Reactions}

\begin{figure}
    \centering
    \includegraphics[width=1.0\linewidth]{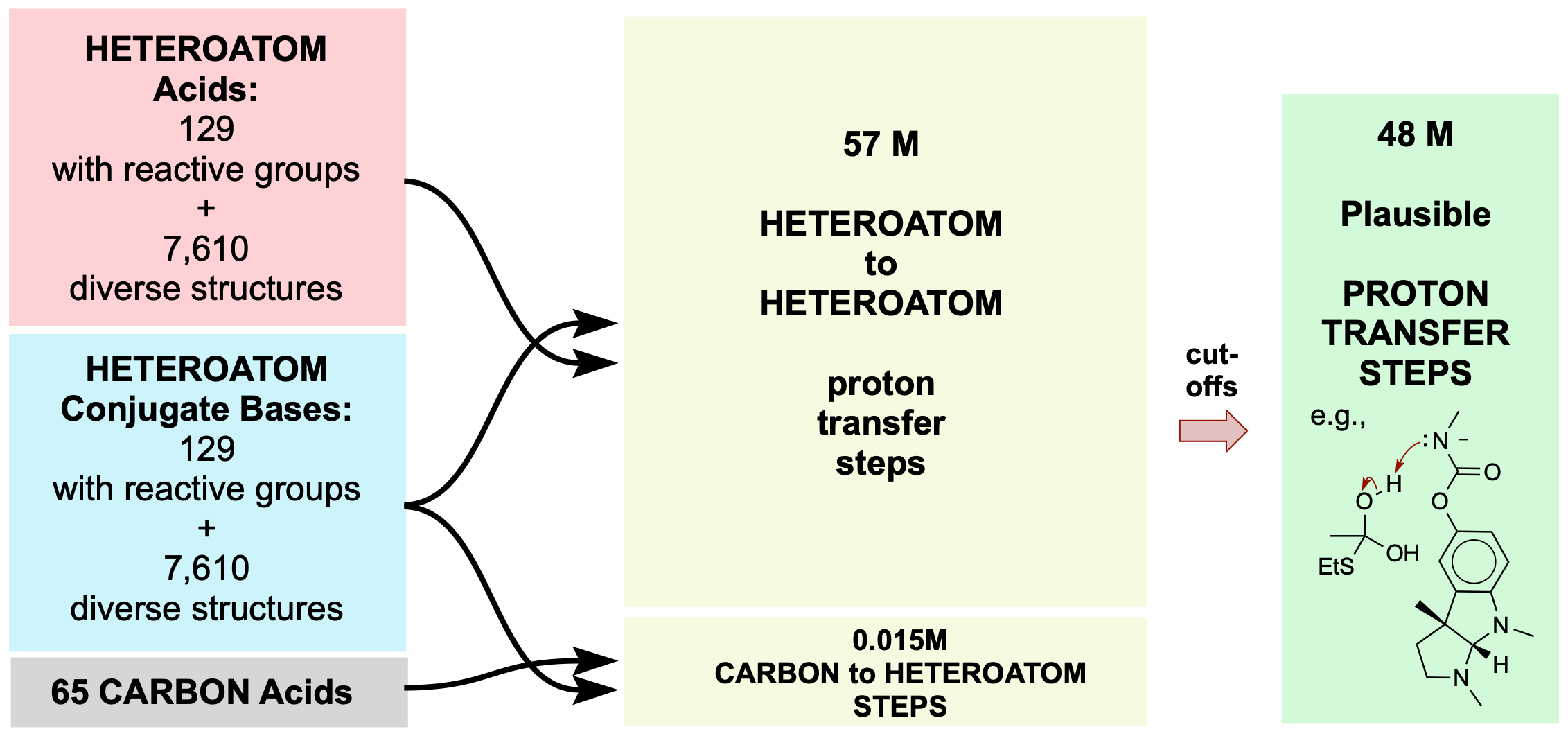}
    \caption{Pipeline for combinatorial reaction generation.}
    \label{fig:combinatorial_pipeling}
\end{figure}

The combinatorial dataset consists of 48,777,226 kinetically plausible proton transfer steps, \cite{Dashuta2025ptdataset} generated combinatorially from over 7,600 acids and 7,600 bases. To construct this dataset, rate constants were estimated from aqueous p\textit{K}\textsubscript{a}s based on the Eigen relationship\cite{crooks1977eigeneq} and conservative boundaries were chosen for inclusion in the dataset. The majority of the acids and conjugate bases were taken from the DataWarrior dataset.\cite{sander2015datawarr} They were structurally diverse with proton donor/acceptor atoms; 98\% had p\textit{K}\textsubscript{a} values in the readily titratable range 0-14. Combinatorial proton transfers were also created using about 100 highly acidic and highly basic heteroatom species from the well-known Reich compilation and Guthrie’s \cite{guthrie1978thioesters, guthrie1993tetrahedral} p\textit{K}\textsubscript{a} estimates for protonated carbonyls and tetrahedral intermediates. Proton transfers between heteroatoms with estimated rate constants $\geq 10^{3}$ M$^{-1}$ s$^{-1}$ --- a conservative boundary --- were included in the dataset. About 15,000 combinatorial proton transfers were generated from carbon acids and heteroatom bases with rate constants estimated using the Eigen-Bernasconi equation.\cite{Bernasconi2005bernasconieq} Steps for proton transfers from carbon with estimated rate constants $\ge$ 10\textsuperscript{-1} M\textsuperscript{-1} s\textsuperscript{-1} were added to the dataset. The pipeline for generating the proton transfer steps can be seen in Figure \ref{fig:combinatorial_pipeling}.

\subsection{Pathway Dataset Curation}

    We chose a well-known intermediate-level organic chemistry textbook \textit{Organic Chemistry, 2nd Ed}., by Clayden, Greeves, and Warren as a source of pedagogically diverse organic transformations. It contains a number of modern transformations missing from introductory textbooks: allylsilane addition, enol silyl ethers, boron aldol reactions, organosulfur chemistry, organophosphorus chemistry, allylsilane reactions, electron transfer reductions, and heterocycle synthesis. Many of the key transformations include arrow-pushing mechanisms for the key steps. 

    We translated 1,187 one-step transformations in the Clayden textbook into entries that included reactant(s), temperature, and product combinations for use in testing product prediction. Many of the entries in the textbook had implied secondary workup steps to generate neutral products. Secondary workups were excluded to ensure that the entries were one-step transformations. All entries include scholarly literature references for the transformation. Many of the entries in the textbook were depicted with generic R substituents, so specific examples were selected from the research literature that best matched the transformation. Each entry has an estimate of the minimum number of elementary mechanistic steps (not included) needed to arrive at the product. We refer to this test set as the 1K Test Set.

    A subset of the 1K Test Set was used to evaluate students from an upper division organic chemistry class at UC Irvine (Chem 125) at the end of the spring quarter of 2023. The Clayden textbook was recommended, but not required, and the course did not follow the structure of the book. We removed from consideration about 400 transformations that require more than one equivalent of a reactant, either due to stoichiometry or need for a mechanistic acid/base, which yielded 800 entries. We then further reduced this student test set to transformations involving seven mechanistic steps or less, leading to 696 entries. Of these, 60 involved chemistry outside of the scope of the class, so they were not considered for the assignment. The total pool of assignable problems was 636 transformations. Each problem consisted of reactant(s), temperature, and a product molecular formula. The justification for including the molecular formula is that such information is readily available from mass spectrometry and investigators are simply trying to match a product mass to a plausible structure. We refer to this subset of the 1K Test Set as the UG Test Set.

The 70 students in the class were each assigned five different randomly chosen transformations from the UG Test Set. The students were asked to propose a product structure in SMILES format consistent with the reactants, temperature, and the molecular formula of the product. For the purposes of grading, students were told that any structure matching the product formula would receive credit.

\begin{figure}[H]
    \centering
    \includegraphics{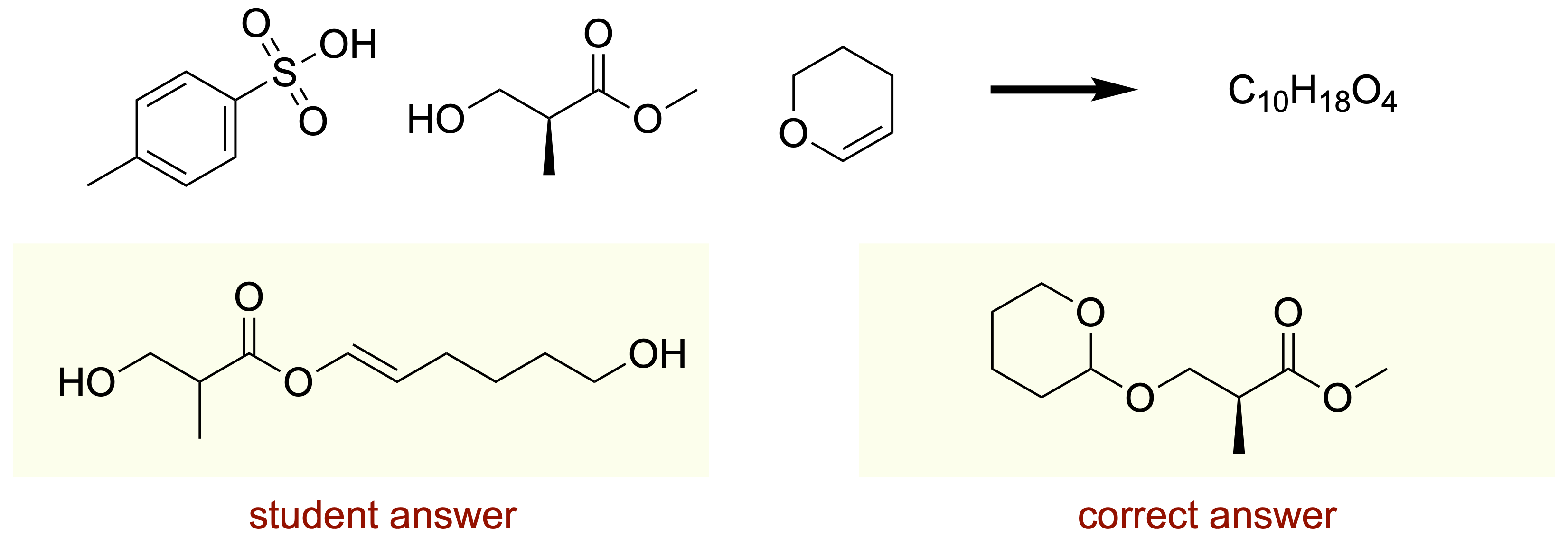}
    \caption{Example of the Transformations Assigned to Students for Product Prediction. PMechRP correctly predicted the target and mechanistic pathway.}
    \label{fig:studentanswer}
\end{figure}

    For 180 of the 350 assigned problems, about half, the student’s product structure was correct. For 21 of the 350 assigned problems, the student’s answer did not match the correct molecular formula. For 149 of the 350 assigned problems, the student’s answer matched the molecular formula but did not match the correct product structure (Figure \ref{fig:studentanswer}). Of these 149 incorrect product structures, 40 were inconsistent with any known transformation and did not appear to arise from a mechanistic analysis. Therefore, 109 of 170 incorrect answers appeared to involve student effort. Of the 289 (180 plus 109) problems attempted, 62\% of the products were correctly identified by students. We refer to this performance as the UG Benchmark.

\subsection{PMechDB Dataset}

\begin{figure}[h]
    \centering
    \includegraphics[scale=0.55]{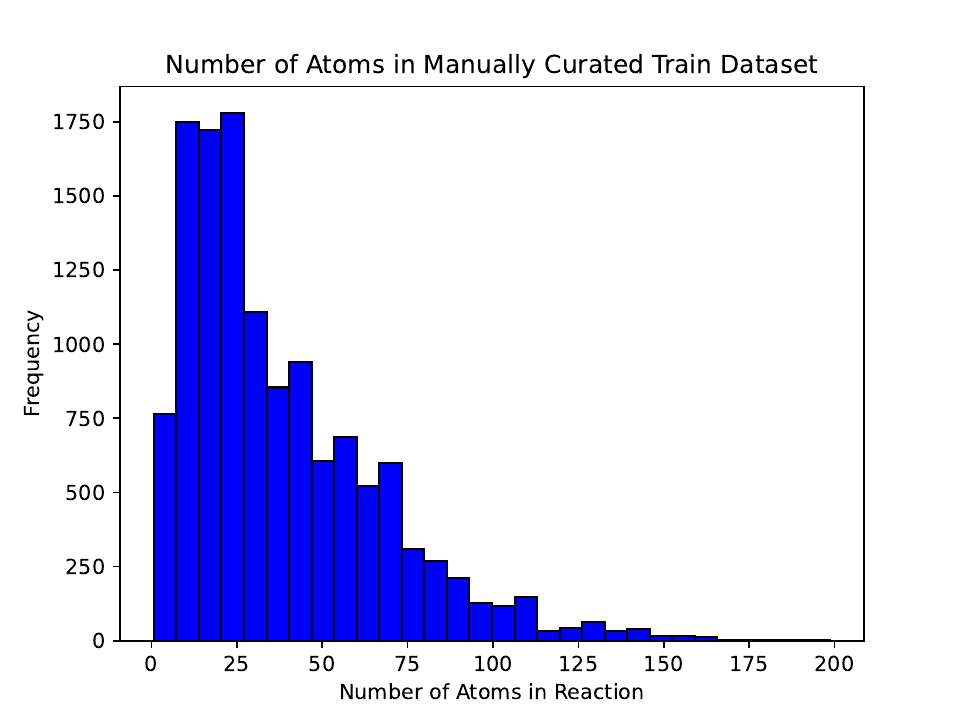}
    \caption{The distribution of the total number of atoms contained in each reaction for the manually curated training dataset.}
    \label{fig:mc_train_len}
\end{figure}

\begin{figure}[h]
    \centering
    \includegraphics[scale=0.55]{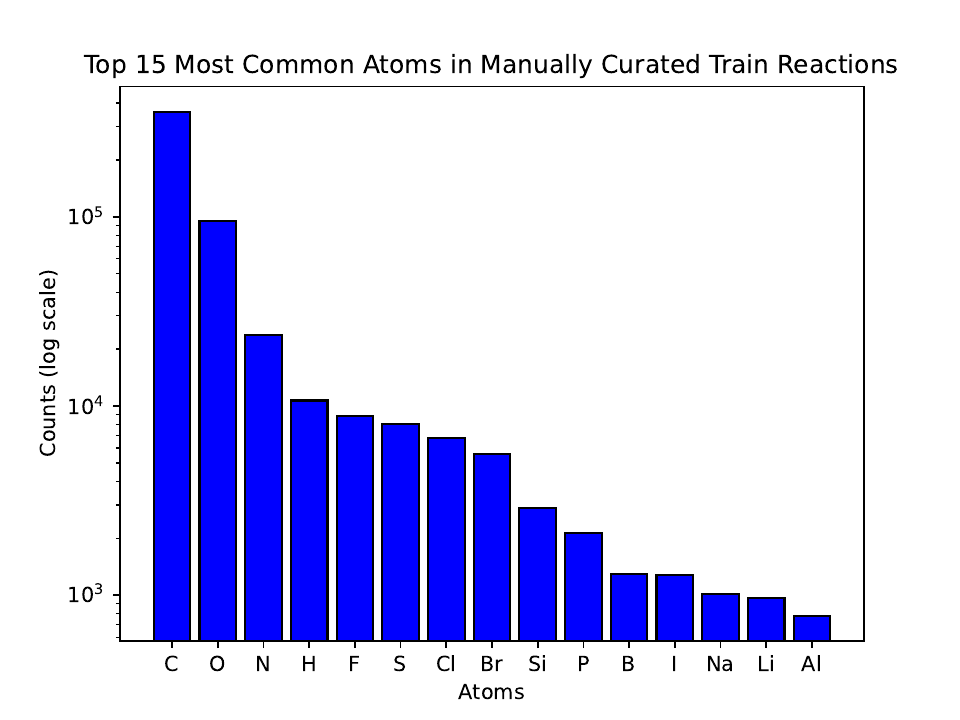}
    \caption{The distribution of atoms for the reactions in the manually curated training dataset.}
    \label{fig:mc_train_atom}
\end{figure}

Here we provide some Figures \ref{fig:mc_train_len}, \ref{fig:mc_train_atom} displaying the the number of atoms and atom types found in the PMechDB dataset \cite{tavakoli2024pmechdb}

\subsection{Model Training}

\subsubsection{Molecular Transformer}

We utilized the text-based reaction predictor, Molecular Transformer \cite{schwaller2019molecular}, which employs a bidirectional encoder and autoregressive decoder coupled with a fully connected network to generate probability distributions over potential tokens. The pre-trained Molecular Transformers underwent training using various versions of the USPTO dataset. We did not separate reactants and reagents, so the model pre-trained using the USPTO\_MIT\_mixed dataset was selected and then fine-tuned on the PMechDB dataset. The hyperparameters can be seen in Table \ref{tab:mt_hyperparams}.

\begin{table}[H]
\centering
\begin{tabular}{ll}
\hline
\textbf{Hyperparameter}         & \textbf{Value} \\
\hline
-train\_steps                    & 500000 \\
-max\_generator\_batches          & 32 \\
-batch\_size                     & 10240 \\
-batch\_type                     & tokens \\
-normalization                  & tokens \\
-max\_grad\_norm                  & 0 \\
-accum\_count                    & 4 \\
-optim                          & adam \\
-adam\_beta1                     & 0.9 \\
-adam\_beta2                     & 0.998 \\
-decay\_method                   & noam \\
-warmup\_steps                   & 8000 \\
-learning\_rate                  & 2 \\
-label\_smoothing                & 0.0 \\
-layers                         & 4 \\
-rnn\_size                       & 256 \\
-word\_vec\_size                  & 256 \\
-encoder\_type                   & transformer \\
-decoder\_type                   & transformer \\
-dropout                        & 0.3 \\
-position\_encoding              & (enabled) \\
-share\_embeddings               & (enabled) \\
\hline
\end{tabular}
\caption{Selected hyperparameters used to train the Molecular Transformer model.}
\label{tab:mt_hyperparams}
\end{table}

\subsubsection{Chemformer}

In addition to the molecular transformer, we adopted the Chemformer model \cite{irwin2022chemformer}, which is another transformer-based reaction predictor. The Chemformer model also employs a bidirectional encoder and autoregressive decoder with a fully connected network to generate probability distributions over potential tokens. The Chemformer model was pre-trained on molecular reconstruction and classification tasks using a dataset of 100M SMILES strings from the ZINC-15 \cite{sterling2015zinc} dataset. Afterwards, the model was fine-tuned onto various downstream tasks, including forward prediction and retrosynthesis. The pre-training substantially improved the model's generalizability and convergence times on downstream tasks, such as USPTO forward prediction, compared to randomly initialized models. We chose to start from the model fine-tuned on USPTO-mixed since reactants and reagents are not separated in the PMechDB dataset. This model was then fine-tuned on the PMechDB dataset for mechanistic-level predictions. The vocabulary of the model was expanded by 66 tokens to account for unseen atoms in the PMechDB dataset. The hyperparameters used can be found in table \ref{tab:chemformer_hyperparams}.

\begin{table}[h!]
\centering
\begin{tabular}{ll}
\hline
\textbf{Hyperparameter}          & \textbf{Value} \\
\hline
task                             & Forward prediction \\
n\_epochs                 & 200 \\
lr                    & 0.001 \\
schedule           & Cyclic \\
batch\_size                       & 16 \\
acc\_batches      & 4 \\
augmentation\_strategy            & All \\
\hline
\end{tabular}
\caption{Hyperparameters used to fine-tune the Chemformer model.}
\label{tab:chemformer_hyperparams}
\end{table}

To improve performance of the chemformer model, we train 5 chemformer models to create an ensemble. In order to add some variance to the models, we vary the augmentation probability. The results of changing the augmentation probability can been seen in table \ref{tab:p_aug}

\begin{table}[htbp]
    \centering
    \caption{Effects of Augmentation Probability on Top-N Accuracy for Chemformer Models}
    \label{tab:p_aug}
    \begin{tabular}{lcccc} 
        \toprule 
        \boldmath{Augmentation Probability} & \textbf{Top-1} & \textbf{Top-3} & \textbf{Top-5} & \textbf{Top-10}\\ 
        \midrule 
        0.0 & 77.1 & 85.6 & 85.9 & 86.1 \\ 
        0.1 & 79.4 & \textbf{87.3} & \textbf{87.5} & \textbf{87.6} \\ 
        0.3 & \textbf{81.2} & 87.0 & 87.2 & 87.4 \\ 
        0.5 & 80.4 & 85.6 & 85.9 & 85.9 \\ 
        0.7 & 79.4 & 84.7 & 84.9 & 84.9 \\ 
        \bottomrule 
    \end{tabular}
\end{table}

We consider the best performing model to be the model with augmentation proability of 0.1, as it achieves the highest top-3, top-5, and top-10 accuracies. To aggregate predictions of the ensemble, we sum the likelihoods from all 5 models for each product. The predictions are then sorted by highest likelihood sum. Ensembling allows the Chemformer models to predict a greater diversity of products, offering smaller increases of around 2\% to top-1 accuracy, but an increase of nearly 4\% to the top-10 accuracy. The impact of varying ensemble sizes on performance for the manually curated dataset is summarized in Table \ref{tab:ensembling}.

\begin{table}[htbp]
    \centering
    \caption{Effects of Ensemble Size on Top-N Accuracy for Chemformer Models}
    \label{tab:ensembling}
    \begin{tabular}{lcccc} 
        \toprule 
        \boldmath{$ensemble size$} & \textbf{Top-1} & \textbf{Top-3} & \textbf{Top-5} & \textbf{Top-10}\\ 
        \midrule 
        2 & 80.3 & 88.3 & 88.6 & 88.9 \\ 
        3 & 81.6 & 90.2 & 90.7 & 90.7 \\ 
        4 & \textbf{82.0} & \textbf{90.7} & 91.2 & 91.2 \\ 
        5 & 81.8 & 90.5 & \textbf{91.4} & \textbf{91.5} \\ 
        \bottomrule 
    \end{tabular}
\end{table}

Pretraining the Chemformer models made a large difference in performance, the effects of pretraining can be seen in Table \ref{tab:pretraining}.

\begin{table}[htbp]
    \centering
    \caption{Effects of Pretraining on Top-N Accuracy of Chemformer Models}
    \label{tab:pretraining}
    \begin{tabular}{lcccc} 
        \toprule 
        \textbf{Model Type} & \textbf{Top-1} & \textbf{Top-3} & \textbf{Top-5} & \textbf{Top-10}\\ 
        \midrule 
        no-pretraining & 43.3 & 56.5 & 57.2 & 57.2 \\ 
        pretrained on zinc and USPTO Mixed & 79.4 & 87.3 & 87.5 & 87.6 \\ 
        \bottomrule 
    \end{tabular}
\end{table}

  The large increase in performance from the pretraining, indicates overlap between the USPTO dataset and the PMechDB dataset. This is in stark contrast to radical mechanisms, which exhibited lower performance when using a pretrained model \cite{tavakoli2024ai}. This suggests that radical reactions are underrepresented in USPTO datasets compared to polar reactions, and that pre-trained transformer models would be expected to have higher performance on polar reactions.

\subsection{T5Chem}

Due to the highly related nature of many chemistry prediction tasks, multitask learning can be used to develop robust models which may demonstrate improved learning efficiency and prediction accuracy. T5Chem is one such model, which leverages multitask learning on a transformer architecture to perform 5 different tasks. The T5Chem multi-task transformer architecture is able to perform forward/backwards prediction, reaction yield prediction, reaction classification, and reagents prediction \cite{lu2022unified}. This architecture was first pretrained with a BERT-like MLM objective on 97 million PubChem molecules. Then, the model was further fine-tuned on 5 different tasks using the USPTO\_500\_MT dataset. We selected this pretrained model, and fine-tuned it on the manually curated and mixed datasets. We trained for 200 epochs, and used the product task type with default hyperparameters.

\subsection{Graph2SMILES}

Lastly, we employed the Graph2SMILES model \cite{tu2022permutation} for reaction prediction, which replaces the traditional sequence-based transformer encoder with a graph encoder to process molecular graphs as inputs. The model uses a Directed Message Passing Neural Network (D-MPNN) to capture local chemical context, followed by a global attention encoder with graph-aware positional embeddings to incorporate topological information and ensure permutation invariance to SMILES formatting, thus eliminating the need for data augmentation. A transformer-based autoregressive decoder then generates the predictions. We select the GAT model which was pretrained on the USPTO\_STEREO dataset. This model was then fine-tuned on the manually curated and mixed datasets. The hyperparameters can be found in table \ref{tab:graph2smiles_hyperparams}.

\begin{table}[H]
\centering
\begin{tabular}{ll}
\hline
\textbf{Hyperparameter}               & \textbf{Value} \\
\hline
MPN\_TYPE                    & dgat \\
MAX\_REL\_POS                & 4 \\
ACCUM\_COUNT                 & 4 \\
ENC\_PE                      & none \\
ENC\_H                       & 256 \\
BATCH\_SIZE                  & 2048 \\
ENC\_EMB\_SCALE              & sqrt \\
MAX\_STEP                    & 200000 \\
ENC\_LAYER                   & 4 \\
BATCH\_TYPE                  & tokens \\
REL\_BUCKETS                 & 11 \\
REL\_POS                     & emb\_only \\
ATTN\_LAYER                  & 6 \\
LR                           & 4 \\
DROPOUT                      & 0.3 \\
REPR\_START                  & smiles \\
REPR\_END                    & smiles \\
\hline
\end{tabular}
\caption{Selected hyperparameters used to train the Graph2SMILES model.}
\label{tab:graph2smiles_hyperparams}
\end{table}

\subsection{Two-Step Prediction}

The two-step prediction model comprises distinct phases. Initially, the model undertakes the task of predicting reactive atoms within the given reaction. Subsequently, these identified reactive sites are paired to formulate potential reaction mechanisms, followed by the application of a ranker model to rank the plausibility of these proposed mechanisms. This architectural design yields highly interpretable predictions, enabling a granular understanding of the model's rationale. When generating predictions, users can discern precisely which atoms are deemed reactive, and they can view the precise arrow-pushing mechanism predicted by the model. From the viewpoint of organic chemists, the two-step architecture offers greater transparency compared to single-step approaches, as the arrow-pushing mechanism provides justification for why the final products were predicted.

\subsubsection{Siamese Architecture}

\begin{figure}
    \centering
    \includegraphics{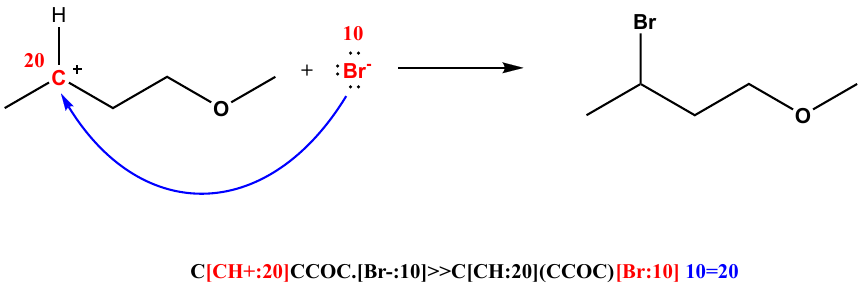}
    \caption{An example of a simple polar elementary step. The electron pushing arrows can be seen in blue, while the source and sink sites are seen in red. The bromine atom labeled 10 is the source atom. The carbon atom labeled 20 is the sink atom. The corresponding SMILES string and arrow codes can be seen below.}
    \label{fig:source_sink_pair}
\end{figure}

The two-step Siamese architecture \cite{fooshee2018deep} comprises three distinct models, each serving a specific function. Initially, two separate reactive atom predictor models are instantiated. One model is specifically trained to predict source atoms, while the other is trained to predict sink atoms. To train the source and sink models, the electron-donating atom from the intermolecular arrow is labeled as the source atom, while the electron-accepting atom is labeled as the sink atom. This labeling process employs the reactive sites identification method as detailed in \cite{fooshee2018deep}. Using the ReactionFP fingerprint, atoms are represented by continuous vectors derived from predefined atomic and graph-topological features, utilizing a neighborhood of size 3. Subsequently, source and sink classifiers are trained to categorize these feature vectors accordingly. Once the trained reactive atom classifiers predict source and sink atoms, these atoms are paired together to enumerate possible arrow-pushing mechanisms via OrbChain\cite{fooshee2018deep, kayala2011learning, tavakoli2024ai}. Afterward, a Siamese architecture is used as a plausibility ranker model, which then ranks the plausibility of each potential mechanism to generate a final set of predictions. A visual representation of the source and sink pair is provided in Figure \ref{fig:source_sink_pair}.

\subsubsection{Reactive Atom Prediction}

A fingerprint of length 6487 is constructed for each atom. This fingerprint consists of 6402 graph-topological features, and 85 hand-crafted physiochemical features. These graph-topological features are extracted using a neighborhood of size 3 with the method described in \cite{fooshee2018deep}, while the physiochemical features are derived from properties such as valence number, electronegativity, aromaticity, atomic number, etc.

The source and sink prediction models are trained using the manually curated and mixed datasets. Each training reaction is processed to extract the atom fingerprints, the atom is given a label 1 if it is reactive, and 0 if it is non-reactive. The final output layer performs a binary classification on a reactive atom. The parameters of the source and sink prediction models can be seen in table \ref{tab:reactive_atom}:

\begin{table}[htbp]
    \centering
    \caption{Source and Sink Model Parameters}
    \label{tab:reactive_atom}
    \begin{tabular}{lccccc} 
        \toprule 
        \textbf{Batch Size} &  \textbf{Num Layers} & \textbf{Layer Dim} & \textbf{Act} & \textbf{Reg} & \textbf{Dropout}\\ 
        \midrule 
        64 & 5 & 512-256-128-164-1 & RELU & L2 & 0.2\\
        \bottomrule 
    \end{tabular}
\end{table}

We assess the performance of the source and sink models on reactive sites identification. The top-N accuracy of the reactive sites identification on the manually curated dataset is presented in Table \ref{tab:reactive_atom_acc}. Reactive site identification is considered correct if both the source and sink atom were correctly identified within the top-N predictions of each model.

\begin{table}[htbp]
    \centering
    \caption{Reactive Atom Classification for Siamese Architecture}
    \label{tab:reactive_atom_acc}
    \begin{tabular}{lcccc} 
        \toprule 
        \textbf{Top-1} & \textbf{Top-3} & \textbf{Top-5} & \textbf{Top-10}\\ 
        \midrule 
        55.4 & 86.2 & 91.1 & 94.3 \\
        \bottomrule 
    \end{tabular}
\end{table}

The source and sink ranking models are able to predict the reactive atoms with relatively high accuracy. Although the reactive atom models are able to filter down the number of potentially reactive atoms significantly, due to the large number of atoms and aromatic structures contained in the polar reactions, enumerating all possible molecular orbital pairs, even from a small set of 10 sources and 10 sinks, creates a vast number of reaction mechanisms for the ranker model to evaluate. 

\subsubsection{Plausibility Ranking}

 The reactionFP fingerprint is extracted using the features explained in \cite{fooshee2018deep}. First, we extract the source and sink features, each of length 6487. Then we compute net change features by using morgan fingerprints of length 2048. We concatenate both the net change features with the source and sink features to create a final reaction fingerprint of length 15022.

The parameters of the ranker model can be seen in table \ref{tab:ranker_params}:

\begin{table}[H]
    \centering
    \caption{Source and Sink Model Parameters}
    \label{tab:ranker_params}
    \begin{tabular}{lcccccc} 
        \toprule 
        \textbf{Batch Size} &  \textbf{Num Layers} & \textbf{Layer Dim} & \textbf{Act} & \textbf{Reg}\\ 
        \midrule 
        200 & 3 & 360-360-1 & Tanh & Dropout (0.5) \\
        \bottomrule 
    \end{tabular}
\end{table}

\end{document}